%% file: main.tex
\documentclass[10pt,twocolumn,letterpaper]{article}
\usepackage[pagenumbers]{cvpr}
\input{preamble}
\definecolor{cvprblue}{rgb}{0.21,0.49,0.74}
\usepackage{placeins}
\usepackage[pagebackref,breaklinks,colorlinks,allcolors=cvprblue]{hyperref}
\usepackage[capitalize]{cleveref}
\usepackage{refcount}
\usepackage{algorithm,algpseudocode,amsmath,amssymb}
\theoremstyle{plain}
\newtheorem{theorem}{Theorem}

\newtheorem*{theorem*}{Theorem}
\newtheorem*{proposition*}{Proposition}
\newcommand{\restatedthmname}{Theorem}
\newtheorem*{restatedthmbody}{\restatedthmname}

\theoremstyle{definition}

\theoremstyle{remark}

\theoremstyle{plain}
\crefname{appendix}{Supp.}{Supp.}
\Crefname{appendix}{Supp.}{Supp.}
\crefname{subappendix}{Supp.}{Supp.}
\Crefname{subappendix}{Supp.}{Supp.}
\crefname{subsubappendix}{Supp.}{Supp.}
\Crefname{subsubappendix}{Supp.}{Supp.}
\newtheoremstyle{boldassumption}
  {\topsep}
  {\topsep}
  {\itshape}
  {}
  {\bfseries}
  {.}
  { .5em}
  {}
\theoremstyle{boldassumption}
\newtheorem{assumption}{Assumption}
\makeatletter
\providecommand{\theHalgorithm}{\arabic{algorithm}}
\providecommand{\theHALG@line}{}
\renewcommand{\theHALG@line}{\theHalgorithm.\arabic{ALG@line}}
\makeatother
\usepackage{pgffor}
\foreach \x in {A,B,C,D,E,F,G,H,I,J,K,L,M,N,O,P,Q,R,S,T,U,V,W,X,Y,Z}{
  \expandafter\xdef\csname c\x\endcsname{\noexpand\mathcal{\x}}
}
\makeatletter
\patchcmd{\@maketitle}{\vskip .375in}{\vskip 0in}{}{\PackageError{main}{title patch failed}{}}
\makeatother
\title{\texttt{CAPTURE}: Minimizing Watch Time for Reliable Temporal Video Grounding}
\author{%
\begin{tabular}[t]{c@{\hspace{1.8em}}c@{\hspace{1.8em}}c}
Ali Asgarov\thanks{Work done during an internship at Amazon.} & Dheeraj Baby & Yifei Tang\\
Virginia Tech, Amazon & Amazon & Amazon\\
{\small\tt aliasgarov@vt.edu, aasgarov@amazon.com} & {\small\tt dheebaby@amazon.com} & {\small\tt yifeitan@amazon.com}
\end{tabular}\\
\noalign{\vskip 0.8em}
\begin{tabular}[t]{c@{\hspace{1.8em}}c}
Rohit Pyati & Chris Thomas\\
Amazon & Virginia Tech\\
{\small\tt pyarohit@amazon.com} & {\small\tt christhomas@vt.edu}
\end{tabular}
}
\begin{document}
\addtocontents{toc}{\protect\setcounter{tocdepth}{-1}}
\raggedbottom
\maketitle
\input{sec/0_abstract}
\input{fig/teaser}
\input{sec/1_intro}
\input{sec/2_related_main}

\input{sec/3_setup}

\input{sec/4_capture}
\input{sec/5_experiments}
\input{sec/6_conclusion}
\clearpage
\addtocontents{toc}{\protect\setcounter{tocdepth}{2}}
\setcounter{dbltopnumber}{3}
\renewcommand{\dbltopfraction}{0.88}
\renewcommand{\dblfloatpagefraction}{0.75}
\renewcommand{\textfraction}{0.08}
\setlength{\dbltextfloatsep}{10pt plus 2pt minus 2pt}
\setlength{\dblfloatsep}{10pt plus 2pt minus 2pt}
\setcounter{theorem}{0}
\renewcommand{\thetheorem}{S\arabic{theorem}}
\renewcommand{\theHtheorem}{S\arabic{theorem}}
\setcounter{algorithm}{0}
\twocolumn[{\begin{center}
{\Large\textbf{\thetitle}}\\[0.5em]
{\Large Supplementary Material}
\end{center}\vspace{1em}}]
\input{sec/S_set}
{
    \small
    \bibliographystyle{ieeenat_fullname}
    \bibliography{main}
}
\end{document}

%% file: preamble.tex
\usepackage[T1]{fontenc}
\usepackage{amsmath}
\usepackage{amssymb}
\usepackage{pifont}

\usepackage{dsfont}
\usepackage{amsthm}
\usepackage{booktabs}
\usepackage{multirow}
\usepackage{graphicx}
\usepackage{adjustbox}
\usepackage{xcolor}
\usepackage{colortbl}
\newcommand{\gray}[1]{{\color{gray}#1}}

\usepackage{algorithm}
\usepackage[noend]{algpseudocode}
\usepackage{microtype}

\newcommand{\best}[1]{\textbf{#1}}
\definecolor{oursblue}{HTML}{2a78d6}
\definecolor{coverorange}{HTML}{eb6834}
\usepackage{cuted}
\usepackage{currfile}
\usepackage{caption}

\makeatletter
\newbox\cvp@labels
\newif\ifcvp@inlabel
\let\cvp@shipoutpicture\@ShipoutPicture
\def\@ShipoutPicture{%
  \global\setbox\cvp@labels\box\@labels
  \if@inlabel\global\cvp@inlabeltrue\else\global\cvp@inlabelfalse\fi
  \global\@inlabelfalse
  \cvp@shipoutpicture
  \global\setbox\@labels\box\cvp@labels
  \ifcvp@inlabel\global\@inlabeltrue\fi}
\makeatother

%% file: sec/0_abstract.tex
\begin{abstract}
Video grounding models typically answer a text query with a single time span, without guaranteeing that it matches the requested moment.
We ask how little of the video a person must watch to find a matching clip with a prescribed probability.
Prior conformal grounding can widen an answer to contain the whole moment, although a viewer only needs footage containing one clip at the required temporal intersection-over-union (tIoU).
We introduce \texttt{CAPTURE}, which uses a frozen grounder's probabilities over candidate spans to find short footage that contains such a clip.
For tIoU of at least 0.5, its exact dynamic program guarantees minimum possible watch time at a given model coverage level.
Conformal calibration on held-out examples then determines how much footage to return to meet the requested success rate.
We establish marginal coverage and a high-probability watch-duration bound relative to an oracle with access to the true conditional distribution.
A matching lower bound shows that the dependence on grounding-model errors is unavoidable in the worst case.
Efficient heuristics retain the coverage guarantee when scoring all spans is impractical.
Experiments on Charades-STA, QVHighlights and ActivityNet Captions
show that \texttt{CAPTURE}, across six grounding backbones,
achieves the prescribed tIoU coverage while reducing average watch
duration by approximately $35\%$ compared with prior methods.
We release our code at \href{https://github.com/aliasgerovs/capture}{\texttt{\upshape https://github.com/aliasgerovs/capture}}.
\end{abstract}

%% file: fig/teaser.tex
\begin{figure}[t]
\centering
\includegraphics[width=\linewidth]{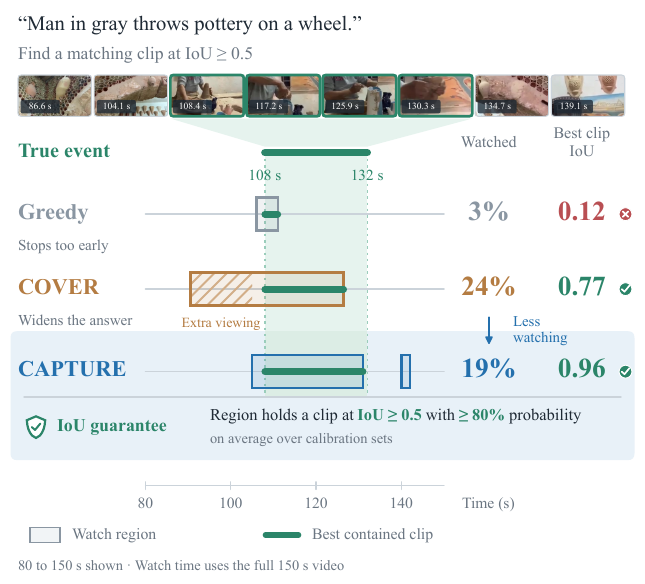}
\caption{Qwen3-VL-8B on QVHighlights.
\textbf{\texttt{CAPTURE} recovers the moment missed by greedy with a better clip than COVER} (IoU 0.96 versus 0.77) while watching less video (19\% versus 24\%).
Hatching marks time watched only by COVER~\cite{mohamed2026cover}, which calibrates ground-truth containment.
Circled marks show hits and misses on this query, and the shield denotes the calibrated IoU guarantee (\cref{eq:coverage_guarantee}).}
\label{fig:teaser}
\end{figure}

%% file: sec/1_intro.tex
\section{Introduction}
\label{sec:intro}

Finding a specific event that matches a text query can require watching long stretches of a video.
Temporal video grounding addresses this challenge by identifying video moments relevant to the query~\cite{gao2017tall,hendricks2017mcn,lei2021qvhighlights}.
However, a predicted interval can miss the event, leaving a human reviewer to search elsewhere in the video.
The prediction alone does not tell the reviewer how much more of the video to watch.
We ask \emph{how little of a video must a person watch to find a matching clip with a guaranteed probability, say 80\%?}

Returning the entire video guarantees that the moment is available to watch, but offers no reduction in viewing effort.
Conformal prediction~\cite{vovk2005algorithmic,shafer2008tutorial} can instead use held-out examples to calibrate how much of the video to return.
COVER~\cite{mohamed2026cover}, to our knowledge the only prior conformal method for video grounding, widens the model's predicted interval at both ends, and calibration decides how far to widen so that the whole moment is contained.
It does not bound how long the widened interval becomes, and a wrong answer must be widened until it covers the whole moment.
Moreover, containing the entire moment is a stronger requirement than giving the viewer one sufficiently accurate match.
A useful guarantee should therefore specify what the returned part of the video must contain and account for how long it takes to watch.

Our key observation is that a viewer does not need the model to find the exact event boundaries, only footage that contains a sufficiently good clip (\cref{fig:teaser}).
The task can therefore be framed as choosing which parts of the video to return, rather than predicting boundaries.
We call the returned parts of the video the \emph{watched region} and its total length the \emph{watch duration}.
The watched region succeeds when it contains a clip whose temporal intersection-over-union (tIoU) with a ground-truth moment is at least a user-specified threshold.
Such a region can be short because it can keep several likely locations and skip the gaps between them.
The grounder assigns a probability to each candidate span on a finite grid.
Adding up the probabilities of the candidate spans that the region matches estimates the probability that it succeeds, its \emph{model-estimated success probability} (\cref{eq:model_utility}).

Building on this observation, we propose \texttt{CAPTURE}.
Instead of widening a single predicted interval as prior work does, \texttt{CAPTURE} first finds the \emph{shortest} watched region with sufficiently high model-estimated success probability.
An exact dynamic program finds this region when the tIoU threshold is at least 0.5, and efficient heuristics approximate it otherwise.
Because the grounder's span probabilities are imperfect, this region is only the model's estimate of the shortest watched region that \emph{succeeds}.
It therefore serves as an initial guess, called the \emph{anchor}, not as the final answer.
Split-conformal calibration on held-out examples then determines how far to expand or shrink the anchor to meet the requested success rate.
We prove that, with high probability, the watch duration of \texttt{CAPTURE} is at most that of an oracle that finds the shortest successful region using the true conditional distribution, at a slightly higher success rate.
A matching lower bound shows that errors in the grounder's span probabilities impose a worst-case watch-duration cost on any calibration method: with a poor grounder, any valid method must watch more.

Across six grounding models on Charades-STA, QVHighlights and ActivityNet Captions, \texttt{CAPTURE} reduces average watch duration by about $35\%$ compared with COVER.
With TimeLens2-8B on QVHighlights at tIoU 0.5 and 80\% requested success rate, it watches 18\% of each video on average, versus 55\% for COVER.
We measure success as how often the watched region contains a matching clip on test videos.
Averaged over 1,000 random calibration--test splits, it falls at most 0.002 below the requested rate in all 108 model--benchmark--request settings.
Further tests cover Charades-TimeLens, ActivityNet-TimeLens and long videos in Charades-STA-Long (\cref{tab:supp:newpools,tab:supp:chlong}).

Our contributions are as follows.
\begin{itemize}
\item We formulate conformal temporal video grounding around a matching-clip guarantee at a user-specified tIoU, with watch duration as the cost.
\item We develop \texttt{CAPTURE}, a split-conformal algorithm that attains a user-specified success rate (marginal coverage). We prove that, with high probability, its watch duration is at most that of an oracle that knows the true conditional distribution, at a slightly higher success rate, and a matching minimax lower bound shows that this watch-duration cost is unavoidable in the worst case. An exact dynamic program computes its anchor for tIoU of at least 0.5, and efficient heuristics approximate it otherwise.
\item We show that \texttt{CAPTURE} attains the requested success rate with less watch duration than COVER across six grounding models and three benchmarks, on short and long videos. Ablations show that cheaper anchors keep the guarantee but watch more, and that one tenth of the calibration videos still meets the requested rate with a small increase in watch duration.
\end{itemize}

%% file: sec/2_related_main.tex
\section{Related work}
\label{sec:related:main}

Temporal grounding models predict intervals~\cite{moon2023qddetr,sun2024trdetr} or generate timestamps with video-language models~\cite{bai2025qwen3vl,zhang2025timelens,zhu2026timelens2,xiaomi2025mimovl}.
Boundary-uncertainty models~\cite{huang2022emb,zeng2025distime} and learned confidence scores~\cite{chen2026confidence} help describe or detect errors, but do not certify how often returned footage contains a matching clip.
COVER~\cite{mohamed2026cover} provides conformal containment guarantees through interval widening or relevance level sets, and also analyzes region size.
\texttt{CAPTURE} changes the target to one clip at the requested tIoU and uses span probabilities to optimize the total footage a viewer must watch.

Conformal prediction~\cite{vovk2005algorithmic,lei2018distribution} calibrates prediction sets, while risk-control methods~\cite{bates2021rcps,angelopoulos2021ltt,angelopoulos2024crc} certify bounds on expected losses.
Conformal language modeling~\cite{quach2024clm} and TRON~\cite{wang2025tron} calibrate sets containing an acceptable answer, and Lu et al.~\cite{lu2026instance} calibrate mask sets in which one mask reaches a requested IoU.
Our finite temporal grid additionally permits exact watch-duration optimization.
Gao et al.~\cite{gao2025volume} study volume-optimal conformal sets, including unions of intervals.
In their regression setting, success is membership of the target in the set.
Here, success requires the union of selected spans to contain a clip matching at the requested tIoU, while cost measures that union.
This gives a different optimization problem.
Extended related work is in \cref{sec:related}.

%% file: sec/3_setup.tex
\section{Setting}
\label{sec:setting}

\paragraph{Temporal video grounding.}
An input $X=(V,q)$ consists of a $T$-second video
$V$ and a text query $q$. Its ground-truth annotation
$Y=\{y_1,\ldots,y_J\}$ contains temporal intervals
$y_j\subseteq[0,T]$ corresponding to relevant
occurrences of the queried event. Charades-STA and
ActivityNet Captions provide one interval per query,
whereas QVHighlights~\cite{lei2021qvhighlights}
may contain multiple. Henceforth, we focus on the
single-ground-truth setting ($J=1$) for theoretical
analysis. Our algorithms also support multiple
moments, which we evaluate experimentally.
Let $P$ denote the joint distribution of $(X,Y)$.

For an input $x$, a grounding model $f$ assigns
probabilities $Q(y\mid x)=p_f(y\mid x)$ to spans
$y$ in a finite candidate pool $\mathcal P$,
built by a fixed rule (e.g., all
integer-second spans).

\paragraph{Coverage and watch duration.}
For temporal intervals $s$ and $y$, their temporal
intersection over union (tIoU) is
$\operatorname{tIoU}(s,y)=|s\cap y|/|s\cup y|$,
where $|\cdot|$ denotes duration in seconds
(or equivalently, units of a uniform temporal grid).

For a prediction set $C\subseteq\mathcal P$, define
its \emph{watched region} $R(C)=\bigcup_{s\in C}s$
and \emph{watch duration} $W(C)=|R(C)|$.
At a prescribed tIoU threshold $\tau\in(0,1]$,
we consider a prediction set successful if at least
one selected span matches a ground-truth annotation:
\begin{equation}
a_\tau(C,Y)=\mathbf{1}\left\{
\exists s\in C,\;y\in Y:
\operatorname{tIoU}(s,y)\ge\tau
\right\},
\label{eq:admit}
\end{equation}
where $\mathbf 1$ is the binary indicator function.
For a single span, we write
$a_\tau(s,Y):=a_\tau(\{s\},Y)$. We say that span \emph{matches} the 
ground-truth $Y$ if $a_\tau(s,Y)=1$

Our goal is to construct prediction sets with
small watch duration while guaranteeing
\emph{marginal coverage} at a user-specified
miscoverage level $\alpha\in(0,1)$:
\begin{equation}
\Pr\bigl(a_\tau(C(X),Y)=1\bigr)\ge1-\alpha.
\label{eq:conf}
\end{equation}
The probability is over $(X,Y)\sim P$ and any
randomness in the procedure, including
calibration data. Without a fixed duration budget,
the watched region adapts to input difficulty.

Although our coverage criterion requires an explicitly
selected span from the prediction set to match the ground truth, 
our ultimate
goal is to find the shortest viewing region containing
a matching clip with the required probability.

These objectives have the same optimal watch duration
when $\mathcal P$ contains all admissible intervals
on a common grid. Indeed, a matching selected span
necessarily lies within the watched region.
Conversely, any grid-aligned viewing region can be
represented by selecting all its admissible
subintervals, without changing its watch duration.
This includes every matching clip within the region,
preserving its success probability and establishing
the equivalence.

\paragraph{Conditional success and optimal watch duration.}
For an input $x$ and prediction set $C$, define
the true conditional success probability as
\begin{equation}
U_P(C\mid x)
=P\bigl(a_\tau(C,Y)=1\mid X=x\bigr).
\label{eq:true_utility}
\end{equation}
Since $P$ is unknown, we use the model-based estimate
\begin{equation}
U_Q(C\mid x)
=\sum_{y\in\mathcal P}Q(y\mid x)\,
a_\tau(C,\{y\}).
\label{eq:model_utility}
\end{equation}

Our ideal benchmark is the minimum watch duration
needed to achieve conditional success probability
at least $r$:
\begin{equation}
\operatorname{OPT}_P(x,r)
=\min_{\substack{C\subseteq\mathcal P\\
U_P(C\mid x)\ge r}}W(C).
\label{eq:opt}
\end{equation}
Although unavailable in practice, this oracle
is a natural benchmark for efficiency. Our objective is to satisfy marginal
coverage~\eqref{eq:conf} while keeping watch
duration close to this conditional benchmark
at a suitable coverage level $r$.

%% file: sec/4_capture.tex
\section{Method and guarantees}
\label{sec:main_results}

\input{fig/method-simple}

We introduce \texttt{CAPTURE}, which combines model-based watch-duration optimization with split-conformal calibration.
We explain its anchor and calibration rule, then establish marginal coverage and a watch-duration bound with asymptotically optimal dependence on model error.

\subsection{\texttt{CAPTURE}: algorithm and design intuition}
\label{sec:capture_algorithm}

\texttt{CAPTURE} first finds the shortest watched region that the model estimates will succeed with probability at least $q_0$ (\cref{fig:method}).
We call $q_0$ the model-estimated success threshold and the resulting prediction set the \emph{anchor}.
The anchor is a model-based starting point, not the final calibrated output.
The initial anchor solves
\begin{equation}
A_0(x)\in
\arg\min_{\substack{C\subseteq\mathcal P\\
U_Q(C\mid x)\ge q_0}}W(C).
\label{eq:oracle}
\end{equation}
Thus, $A_0(x)$ minimizes watch duration among prediction sets with model-estimated success probability at least $q_0$.

For our analysis, we make $A_0(x)$ \textbf{inclusion-minimal}.
We examine its spans in increasing order of model probability and remove any span whose deletion keeps the model-estimated success probability at least $q_0$.
This step cannot increase watch duration.
We write the resulting anchor as $A(x):=\operatorname{Anchor}_Q(x;q_0)$, as in Line 2 of Algorithm~\ref{alg:capture_calibration}.
The choice of $q_0$ affects watch duration, but marginal coverage holds for any $q_0$ fixed before calibration.

Because the model probabilities can be wrong, split-conformal calibration decides whether to shrink or expand the anchor.
For each input $x$, we order all $N=|\mathcal P|$ candidate spans as $(s_1,\ldots,s_N)$.
We place the anchor spans first, then the remaining spans, and sort each group by decreasing model probability with fixed tie-breaking.
We denote this ordering by $\operatorname{Order}(x,A(x))$, as in Line 3 of Algorithm~\ref{alg:capture_calibration}.
For each prefix $C_j(x)=\{s_1,\ldots,s_j\}$, we define
\begin{equation}
u_j(x)=U_Q(C_j(x)\mid x),\qquad j=0,\ldots,N,
\label{eq:prefix_utility}
\end{equation}
where $C_0(x)=\emptyset$.
Thus, $u_j(x)$ is the model-estimated success probability of the first $j$ spans.
These prefixes are nested, and $u_j(x)$ cannot decrease as $j$ grows.

For calibration example $i$, let $j_i^\star$ be the first prefix whose watched region contains a matching clip.
Its nonconformity score $Z_i$ is the model-estimated success probability immediately before this first success, as in Line 7 of Algorithm~\ref{alg:capture_calibration}.
A large score means that more model probability was needed before the first successful prefix, and any prefix above $Z_i$ includes it.
The conformal quantile $\hat q$ in Line 9 determines the smallest test prefix with model-estimated success probability above $\hat q$, as in Line 6 of Algorithm~\ref{alg:capture_prediction}.
Algorithms~\ref{alg:capture_calibration} and \ref{alg:capture_prediction} give these calibration and test steps.

\begin{algorithm}[t]
\caption{\texttt{CAPTURE}: Conformal calibration}
\label{alg:capture_calibration}
\begin{algorithmic}[1]
\Require Calibration data
$\mathcal D_n=\{(X_i,Y_i)\}_{i=1}^n$, Grounding model $Q$, candidate pool $\mathcal P$,
    tIoU threshold $\tau$, model-estimated success threshold $q_0$,
target miscoverage $\alpha$
\For{$i=1,\ldots,n$}
    \State $A_i\gets\operatorname{Anchor}_Q(X_i;q_0)$
    \State $(s_1,\ldots,s_N)\gets
           \operatorname{Order}(X_i,A_i)$
    \State Compute $u_j(X_i)$ for $j=0,\ldots,N$
    \State $C_j(X_i) \gets \{s_1,\ldots,s_j \}$
    \State $j_i^\star\gets
           \min\{j:a_\tau(C_j(X_i),Y_i)=1\}$
    \State $Z_i\gets u_{j_i^\star-1}(X_i)$ \Comment{See Eq.~\eqref{eq:prefix_utility}}
\EndFor
\State $k_n\gets\lceil(n+1)(1-\alpha)\rceil$
\State $\hat q\gets Z_{(k_n)}$ if $k_n\le n$,
       else $\hat q\gets1$
\State \Return Calibrated threshold $\hat q$
\end{algorithmic}
\end{algorithm}

The anchor-first ordering also controls watch duration.
Let $\widehat C(x)$ be the final prediction set returned by Algorithm~\ref{alg:capture_prediction}.
Since the complete anchor has model-estimated success at least $q_0$, the selected prefix stays inside it whenever $\hat q<q_0$:
\begin{equation}
\widehat C(x)\subseteq A(x)
\quad\Longrightarrow\quad
W(\widehat C(x))\le W(A(x)).
\label{eq:anchor_containment}
\end{equation}
Thus, the calibrated watched region is no longer than the anchor.
If $\hat q<q_0$, it can stop inside the anchor.
If $\hat q\ge q_0$, it reaches the whole anchor before adding spans outside it.

Inclusion-minimality makes every proper anchor prefix have model-estimated success probability below $q_0$.
Therefore, a calibration example covered by its anchor has a nonconformity score below $q_0$.
If at least $k_n$ calibration anchors succeed, with $k_n$ defined in Line 8, then $\hat q<q_0$ and~\eqref{eq:anchor_containment} applies.
Assumption~\ref{ass:model} and a sufficiently large $q_0$ make this event hold with high probability.

\begin{algorithm}[t]
\caption{\texttt{CAPTURE}: Prediction set construction}
\label{alg:capture_prediction}
\begin{algorithmic}[1]
\Require Test input $x$, model $Q$,
    candidate pool $\mathcal P$, model-estimated success threshold $q_0$,
calibrated threshold $\hat q$
\State $A\gets\operatorname{Anchor}_Q(x;q_0)$
\State $(s_1,\ldots,s_N)\gets\operatorname{Order}(x,A)$
\If{$\hat q=1$}
    \State \Return $\widehat C(x)=\mathcal P$
    \Statex \hspace{\algorithmicindent}\emph{(all candidate subclips from the entire video)}
\EndIf
\State Compute $u_j(x)$ for $j=0,\ldots,N$
\State $\hat j(x)\gets\min\{j:u_j(x)>\hat q\}$
\State \Return $\widehat C(x)
       =\{s_1,\ldots,s_{\hat j(x)}\}$
\end{algorithmic}
\end{algorithm}

\subsection{Analysis}
\label{sec:capture_theory}

We prove distribution-free marginal coverage and a high-probability watch-duration bound relative to~\eqref{eq:opt}.
We use the following assumptions.

\begin{assumption}[Sampling]
\label{ass:xchange}
The calibration examples $(X_i,Y_i)_{i=1}^n$ and
test example $(X_{\mathrm{test}},Y_{\mathrm{test}})$
are drawn i.i.d.\ from $P$. The model $Q$, candidate
pool $\mathcal P$, and tie-breaking rules are fixed
independently of the calibration data.
\end{assumption}

\begin{assumption}[Model misspecification]
\label{ass:model}
For each input $x$, define the maximum
over- and underestimation errors as
\begin{align}
e_+(x)
&=\sup_{C\subseteq\mathcal P}
\bigl(U_Q(C\mid x)-U_P(C\mid x)\bigr)_+,
\label{eq:overestimation_error}\\
e_-(x)
&=\sup_{C\subseteq\mathcal P}
\bigl(U_P(C\mid x)-U_Q(C\mid x)\bigr)_+.
\label{eq:underestimation_error}
\end{align}
Here $(x)_+ := \max \{x ,0 \}$. There exist $\epsilon_A,\epsilon_L\ge0$ and
$\beta\in[0,1)$ such that
\begin{equation}
\begin{aligned}
\mathbb E_X[e_+(X)]&\le\epsilon_A,\\
\Pr_X(e_-(X)\le\epsilon_L)&\ge1-\beta.
\end{aligned}
\label{eq:model_assumption}
\end{equation}
\end{assumption}

\begin{assumption}[Realizability]
\label{ass:real}
For $P$-almost every $(X,Y)$, there exists a
candidate span $s\in\mathcal P$ satisfying
$a_\tau(s,Y)=1$.
\end{assumption}

\begin{assumption}[Optimization oracle]
\label{ass:oracle}
We have access to an exact oracle for
Eq.~\eqref{eq:oracle}, followed by the
inclusion-minimality post-processing described
in~\cref{sec:capture_algorithm}.
\end{assumption}

Assumption~\ref{ass:xchange} is the standard split-conformal condition that calibration and test examples are i.i.d.\ and all design choices are fixed before calibration.
In Assumption~\ref{ass:model}, $e_+(x)$ is the largest model overestimate of success, while $e_-(x)$ is the largest underestimate.
The value $\epsilon_A$ bounds average overestimation, while $\epsilon_L$ bounds underestimation on at least a $1-\beta$ fraction of inputs.
Overestimation can make the anchor less reliable than predicted, while underestimation can hide shorter reliable regions.
Average control of overestimation is enough to show that calibration stays within the anchor, while the oracle comparison needs high-probability control of underestimation.
Unlike the two-sided high-probability bounds in~\cite{gao2025volume}, we control overestimation only in expectation.

Assumption~\ref{ass:real} requires at least one matching candidate span, which holds if the pool includes the annotated span.
Assumption~\ref{ass:oracle} separates the statistical result from anchor computation.
We give an exact $O(T^3)$ dynamic program for $\tau\ge0.5$ when selected segments are at least one second apart.
For $\tau<0.5$, or when scoring all spans is impractical, we use efficient heuristics described in the supplement.
We now state the coverage and watch-duration guarantees.

\begin{theorem}[Coverage and watch-duration guarantee]
\label{thm:main}
Suppose Assumptions~\ref{ass:xchange} and
\ref{ass:real} hold. For any $q_0\in[0,1]$
fixed independently of calibration data,
the prediction set $\widehat C(X_{\mathrm{test}})$
returned by \texttt{CAPTURE} satisfies
\begin{equation}
\Pr\bigl(
a_\tau(\widehat C(X_{\mathrm{test}}),
Y_{\mathrm{test}})=1
\bigr)\ge1-\alpha.
\label{eq:coverage_guarantee}
\end{equation}

Furthermore, suppose Assumptions~\ref{ass:model}
and \ref{ass:oracle} also hold. Let
$\delta\in(0,1)$ and
$k_n=\lceil(n+1)(1-\alpha)\rceil\le n$.
If $q_0$ satisfies
\begin{equation}
q_0\ge
\frac{k_n}{n}+\epsilon_A+
\sqrt{\frac{\log(1/\delta)}{2n}},
\qquad q_0+\epsilon_L\le1,
\label{eq:q0_condition}
\end{equation}
then
\begin{equation}
\Pr\bigl(
W(\widehat C(X_{\mathrm{test}}))
\le \operatorname{OPT}_P(
X_{\mathrm{test}},q_0+\epsilon_L)
\bigr)
\ge1-\delta-\beta,
\label{eq:watch_duration_guarantee}
\end{equation}
where the probabilities are taken over randomness in calibration and test
data.
\end{theorem}

The first part of Theorem~\ref{thm:main} gives marginal coverage for any $q_0$ fixed before calibration, regardless of model accuracy.
The second part bounds watch duration by $\operatorname{OPT}_P(X_{\mathrm{test}},q_0+\epsilon_L)$ with probability at least $1-\delta-\beta$.
The condition in~\eqref{eq:q0_condition} accounts for the conformal quantile, average overestimation and finite calibration data, while keeping $q_0+\epsilon_L\le1$.
This condition is sufficient but not necessary, so we tune $q_0$ on separate validation data without affecting marginal coverage.

For the smallest allowed $q_0$ in~\eqref{eq:q0_condition}, the oracle comparison uses coverage level
\begin{equation}
r_n=\frac{k_n}{n}+\epsilon_A+\epsilon_L+
\sqrt{\frac{\log(1/\delta)}{2n}}.
\label{eq:coverage_relaxation}
\end{equation}
Thus, $r_n$ is the requested success rate plus model-error and finite-sample terms.
Since $k_n/n=1-\alpha+O(1/n)$, we have $r_n=1-\alpha+\epsilon_A+\epsilon_L+o(1)$ whenever $\log(1/\delta)=o(n)$.
If $\epsilon_A+\epsilon_L<\alpha$ and $\beta=0$, we can let $\delta\to0$ slowly enough that the comparison holds with probability tending to one.

Model errors therefore raise the oracle coverage level needed for this comparison.
The following lower bound shows that this cost cannot be removed uniformly, even with arbitrarily many calibration examples.
It applies to all marginally valid procedures, including methods outside conformal prediction.

Let $\mathfrak F_0(\epsilon_A,\epsilon_L)$ contain the realizable distribution--model pairs satisfying Assumption~\ref{ass:model} with $\beta=0$.
Let $\mathfrak A_n$ contain all possibly randomized algorithms that use $n$ i.i.d.\ labeled examples, $Q$ and a test input to guarantee marginal coverage at least $1-\alpha$ over this class.
We use all integer-endpoint intervals on a fixed discretized timeline.

\begin{theorem}[Minimax sharpness of coverage relaxation]
\label{thm:lb_minimax_sharpness}
Fix $\tau,\alpha\in(0,1)$ and
$\epsilon_A,\epsilon_L\ge0$ with
$\epsilon_A+\epsilon_L<\alpha$. Define
\begin{equation}
r^\star=1-\alpha+\epsilon_A+\epsilon_L.
\label{eq:critical_level}
\end{equation}
There exist a sufficiently large fixed temporal
grid length $T$ and a constant $K_0>1$ such that,
for every fixed $r\in(0,r^\star)$, there is a
constant $c_r>0$, independent of $n$, satisfying
\begin{equation}
\begin{aligned}
&\liminf_{n\to\infty}
\inf_{\mathcal A_n\in\mathfrak A_n}
\sup_{(P,Q)\in\mathfrak F_0}
\\
&\quad\Pr\left(
\frac{W(\widehat C_n(X))}
{\operatorname{OPT}_P(X,r)}
\ge K
\right)\ge c_r
\end{aligned}
\label{eq:minimax_lower_bound}
\end{equation}
for every $K\in(1,K_0]$. The conclusion also
holds for any sequence $K_n>1$ with $K_n\to1$.
The hard instances can be chosen with both
$P(Y\mid X=x)$ and $Q(\cdot\mid x)$ varying
nontrivially with $x$.
\end{theorem}

The lower bound also holds for $\beta>0$, which allows a larger distribution class.
No uniformly valid method can match the oracle below level $1-\alpha+\epsilon_A+\epsilon_L$ with probability tending to one.
Model errors can hide which inputs allow short watched regions, so this limit applies beyond conformal calibration.
The two theorems thus match asymptotically in their dependence on model error.
We study the extra watch duration from finite calibration data in \cref{sec:exp:ablations}.

%% file: fig/method-simple.tex
\begin{figure*}[t]
\centering
\includegraphics[width=\textwidth]{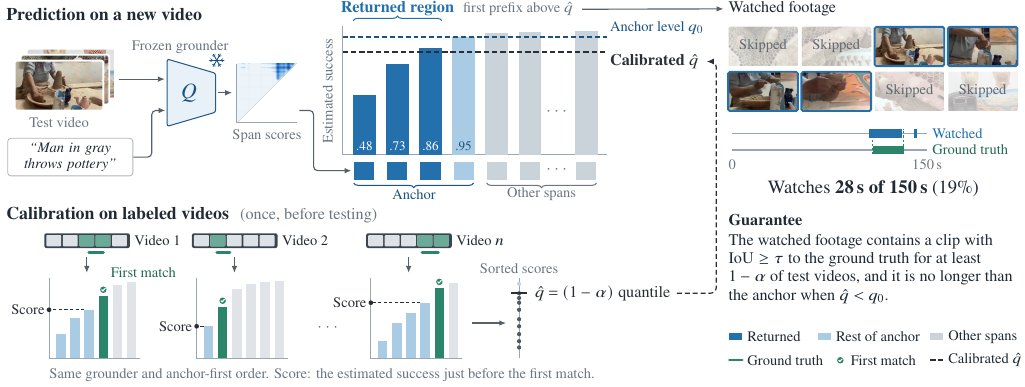}
\caption{\textbf{\texttt{CAPTURE} stops watching once the model is confident enough, and calibration decides what counts as enough.}
Top: a frozen grounder scores every span, and \texttt{CAPTURE} grows the watched region anchor first until its estimated success passes the calibrated threshold $\hat q$; here it watches 28 of 150 seconds (\cref{alg:capture_prediction}).
Bottom: $\hat q$ is set once from labeled videos, as the $(1-\alpha)$ quantile of the estimated success just before each video's first match (\cref{alg:capture_calibration}).
\textbf{The watched footage then contains a clip with IoU at least $\tau$ for at least $1-\alpha$ of test videos.}}
\label{fig:method}
\end{figure*}

%% file: sec/5_experiments.tex
\section{Experiments}
\label{sec:exp}

\input{tables/set_main}

We test whether the watched region contains a matching clip at the requested success rate.
We also measure how much of the video a person must watch to obtain this guarantee.

\par\addvspace{2ex plus .5ex minus .2ex}\noindent\textbf{Benchmarks and models.}
We evaluate three general video-language models and three grounding-tuned models on Charades-STA~\cite{gao2017tall,sigurdsson2016hollywood}, QVHighlights~\cite{lei2021qvhighlights} and ActivityNet Captions~\cite{krishna2017dense}.
\Cref{tab:main} compares \texttt{CAPTURE}, its greedy-anchor variant and COVER for these six grounding models.
Each model outputs a start time and an end time.
We use its output probabilities to compute $Q(y\mid x)$ for every whole-second candidate span $y$, as described in \cref{sec:supp:impl:scoring}.
The \emph{greedy answer} is the span produced by greedy decoding.

\par\addvspace{2ex plus .5ex minus .2ex}\noindent\textbf{Evaluation protocol.}
We choose the model-estimated success threshold $q_0$ on validation set $\mathcal D_{\mathrm{fit}}$, formed from 30\% of each benchmark's COVER calibration and test pools~\cite{mohamed2026cover}.
We choose $q_0$ to give the smallest mean watch duration over all validation queries under \cref{alg:capture_calibration,alg:capture_prediction}.
We then fix $q_0$ and run 1,000 trials, each splitting the remaining videos into 40\% for calibration and 60\% for testing.
Since $q_0$ is fixed before calibration, \cref{thm:main} still gives marginal coverage.
In each trial, the calibration videos set $\hat q$ (\cref{alg:capture_calibration}), and the test videos are scored with \cref{alg:capture_prediction}.
On QVHighlights, clips from the same source video count as one video.
Algorithm~\ref{alg:capture_calibration} uses one randomly selected query per calibration video, so examples are independent across videos as required by Assumption~\ref{ass:xchange}.

For long videos, evaluating every candidate span during anchor construction is expensive.
We keep the 12,000 spans with the highest model probabilities and renormalize their probabilities to sum to 1.
We construct the anchor from this retained pool.
If the retained pool contains no matching clip for a calibration example, its score is 1.
A threshold below 1 returns the calibrated prefix of the retained pool, while threshold 1 returns all admissible subclips in the entire video.
This fallback preserves marginal coverage, but the pruned optimization is not exactly watch-duration optimal.

A request pairs a tIoU threshold $\tau\in\{0.5,0.7\}$ with a success rate $1-\alpha$ of 70\%, 80\% or 90\%.
Hit is the fraction of examples where the watched region contains a matching clip at the requested tIoU, estimating the success probability in~\eqref{eq:conf}.
Watch is the watch duration $W(\widehat C(X))$ over the video length $T$.
Segments (Seg.) counts the separate parts of the watched region.
We average each metric over a video's queries, then over videos, giving each video equal weight.

\input{fig/model-watch}

\par\addvspace{2ex plus .5ex minus .2ex}\noindent\textbf{Comparison methods.}
COVER~\cite{mohamed2026cover} expands both ends of the greedy answer by a calibrated multiple of its length.
The multiple gives whole-moment containment with success rate $1-\alpha$.
The other methods keep \texttt{CAPTURE}'s tIoU target, trials and split-conformal calibration, but replace its exact anchor with a cheaper region.
This replacement keeps marginal coverage when exact optimization is too slow.

The greedy-anchor method widens the greedy answer about its center using our tIoU loss.
The MBR-anchor method similarly widens a minimum Bayes risk (MBR) answer chosen from the span probabilities.
It tests whether a better single answer is enough to reduce watch duration.
Per-second thresholding selects likely times along the video and can return several separated segments.
It tests complete-span probabilities against probabilities assigned separately along the timeline.
The positional prior uses no grounding-model probabilities and tests whether a moment's typical location is enough.
\Cref{tab:methods,tab:supp:methods} also compare fixed regions and calibrated choices between greedy and the whole video.
Implementation details are in \cref{sec:supp:impl:baselines}.

\begingroup
\clubpenalty=10000
\refstepcounter{subsection}
\makeatletter
\def\@currentlabelname{Validity}
\makeatother
\par\addvspace{2ex plus .5ex minus .2ex}\noindent\textbf{\thesubsection.\ Validity.}
\label{sec:exp:validity}
Across 108 model--benchmark--request combinations, mean Hit over 1,000 trials lies between 0.002 below and 0.016 above the request (\cref{tab:main}).
This agrees with the marginal guarantee of \cref{thm:main}, which averages over calibration sets and a new video with a randomly selected query.
A single trial can still fall below the target.
Its success rate depends on which calibration videos are drawn, and averaging over 1,000 trials estimates the marginal rate.
A method can also raise Hit with more footage, so Watch measures the cost at the same requested tIoU and success rate.
\par\endgroup

\refstepcounter{subsection}
\makeatletter
\def\@currentlabelname{Main results}
\makeatother
\par\addvspace{2ex plus .5ex minus .2ex}\noindent\textbf{\thesubsection.\ Main results.}
\label{sec:exp:main}
Across seven Qwen3-VL and TimeLens2 variants (2B--32B), \texttt{CAPTURE} watches less than COVER in all 126 settings and less than its greedy-anchor variant in 120 (\cref{tab:supp:size}).

The calibrated rule adapts watch duration to each query and usually returns one segment (\cref{fig:watch-distribution}).
A test prefix grows until its model-estimated success probability $u_j(x)$ exceeds $\hat q$, so concentrated span probabilities stop it early and uncertain ones need more spans.
A higher success rate must cover more video--query pairs, while a higher tIoU needs a closer boundary match.
Either can need more footage, but neither gives every query the same watch duration.
The mean mixes short regions with difficult queries, so an average reduction need not hold for every query.

\refstepcounter{subsection}
\makeatletter
\def\@currentlabelname{Comparison with other regions}
\makeatother
\par\addvspace{2ex plus .5ex minus .2ex}\noindent\textbf{\thesubsection.\ Comparison with other regions.}
\label{sec:exp:compare}
\texttt{CAPTURE} watches less than COVER and its greedy-anchor variant in 108 and 98 settings, respectively (\cref{tab:main,tab:methods,fig:model-watch}).
The MBR-anchor variant watches more than \texttt{CAPTURE} in 79 settings, and the positional prior watches more in every setting.
Single-answer anchors remain competitive in some cases.
When the span probabilities concentrate around one location, the shortest watched region is one segment, close to a widened single answer.
The benefit of selecting several spans depends on the span probabilities and the request.

Part of the gain over COVER comes from requiring one matching clip instead of containing every ground-truth moment.
The greedy-anchor and MBR-anchor comparisons keep this goal and calibration fixed, isolating the benefit of choosing a region from span probabilities.
\input{tables/set_methods}
\input{fig/qualitative}

On QVHighlights at tIoU 0.5 and 80\% success rate, TimeLens2-8B watches 0.18 versus 0.22 with the greedy anchor.
When two separated parts are plausible, expanding one answer to reach both also includes the gap.
\texttt{CAPTURE} can keep both probability peaks while leaving that gap unwatched (\cref{fig:teaser,fig:qualitative}).
In \cref{fig:qualitative}, greedy selects only the earlier peak and misses the moment, while the two \texttt{CAPTURE} segments contain a matching clip.

\refstepcounter{subsection}
\makeatletter
\def\@currentlabelname{Calibration size}
\makeatother
\par\addvspace{2ex plus .5ex minus .2ex}\noindent\textbf{\thesubsection.\ Calibration size.}
\label{sec:exp:ablations}
We keep the grounding model and validation choice of $q_0$ fixed, changing only the number of calibration videos.
With one tenth of calibration videos (13 on QVHighlights), mean success at tIoU 0.5 stays above the 80\% request in all 18 model--benchmark pairs.
TimeLens2-8B's QVHighlights watch duration rises from 0.18 to 0.23 (\cref{tab:supp:calsize}).
Fewer scores can make the conformal quantile return a longer region.
With $n=13$ and $\alpha=0.2$, Line 8 of \cref{alg:capture_calibration} gives $k_n=12$, so $\hat q$ is the second-largest of 13 scores.
\Cref{thm:main} shows the same effect, since its finite-sample term $\sqrt{\log(1/\delta)/(2n)}$ in~\eqref{eq:q0_condition} grows as $n$ falls and raises the oracle level $r_n$ in~\eqref{eq:coverage_relaxation}.
This extra watch duration is the cost of less calibration data, while the requested tIoU and success rate remain met.

%% file: tables/set_main.tex
\begin{table*}[t]
\caption{\textbf{\texttt{CAPTURE} meets requested coverage with less watch time across models and benchmarks.}
Top shows \texttt{CAPTURE} at six requests. Higher coverage or tIoU targets require longer regions, usually one segment.
Hit measures how often the footage contains a ground-truth match at the requested tIoU.
Watch divides watch duration by video length, $W(C(X))/T$. Seg.\ counts separate watched segments.
Bottom compares Watch at tIoU 0.5 and 80\% success. Greedy anchor is \texttt{CAPTURE} with the greedy answer, widened about its center, in place of the exact anchor.
COVER instead certifies whole-moment containment, averaged over query-level calibration sets.
\textbf{Bold} marks least watch duration per benchmark and request above, and per model--benchmark pair below.}
\label{tab:main}
\centering
\scriptsize
\setlength{\tabcolsep}{2.4pt}
\begin{adjustbox}{max width=\textwidth}
\begin{tabular}{@{}ll ccc ccc ccc ccc cc@{}}
\toprule
& & \multicolumn{3}{c}{IoU\,$\ge$\,0.5 for 70\%} & \multicolumn{3}{c}{IoU\,$\ge$\,0.5 for 80\%} & \multicolumn{3}{c}{IoU\,$\ge$\,0.7 for 70\%} & \multicolumn{3}{c}{IoU\,$\ge$\,0.7 for 80\%} & \multicolumn{1}{c}{0.5, 90\%} & \multicolumn{1}{c}{0.7, 90\%} \\
\cmidrule(lr){3-5}\cmidrule(lr){6-8}\cmidrule(lr){9-11}\cmidrule(lr){12-14}\cmidrule(lr){15-15}\cmidrule(lr){16-16}
\textbf{Benchmark} & \textbf{Model} & Hit & Watch$\downarrow$ & Seg. & Hit & Watch$\downarrow$ & Seg. & Hit & Watch$\downarrow$ & Seg. & Hit & Watch$\downarrow$ & Seg. & Watch$\downarrow$ & Watch$\downarrow$ \\
\midrule
Charades-STA & Qwen3-VL-4B~\cite{bai2025qwen3vl} & 0.70 & 0.27 & 1.16 & 0.80 & 0.34 & 1.22 & 0.70 & 0.37 & 1.14 & 0.80 & \best{0.42} & 1.15 & 0.50 & \best{0.55} \\
 & Qwen3-VL-8B~\cite{bai2025qwen3vl} & 0.70 & \best{0.26} & 1.12 & 0.80 & \best{0.33} & 1.14 & 0.70 & \best{0.36} & 1.09 & 0.80 & \best{0.42} & 1.10 & \best{0.47} & \best{0.55} \\
 & MiMo-VL-7B~\cite{xiaomi2025mimovl} & 0.70 & 0.33 & 1.12 & 0.80 & 0.43 & 1.10 & 0.70 & 0.44 & 1.06 & 0.80 & 0.54 & 1.04 & 0.54 & 0.66 \\
\cmidrule(lr){2-16}
 & TimeLens-8B~\cite{zhang2025timelens} & 0.70 & 0.30 & 1.07 & 0.80 & 0.39 & 1.04 & 0.70 & 0.39 & 1.04 & 0.80 & 0.47 & 1.03 & 0.48 & 0.56 \\
 & TimeLens2-4B~\cite{zhu2026timelens2} & 0.70 & 0.29 & 1.04 & 0.80 & 0.38 & 1.05 & 0.70 & 0.39 & 1.01 & 0.80 & 0.46 & 1.02 & 0.51 & 0.60 \\
 & TimeLens2-8B~\cite{zhu2026timelens2} & 0.70 & 0.31 & 1.04 & 0.80 & 0.38 & 1.05 & 0.70 & 0.39 & 1.02 & 0.80 & 0.46 & 1.01 & 0.51 & 0.60 \\
\midrule
QVHighlights & Qwen3-VL-4B~\cite{bai2025qwen3vl} & 0.70 & 0.18 & 1.16 & 0.80 & 0.23 & 1.27 & 0.70 & 0.25 & 1.13 & 0.81 & 0.32 & 1.21 & 0.39 & 0.44 \\
 & Qwen3-VL-8B~\cite{bai2025qwen3vl} & 0.70 & 0.17 & 1.23 & 0.81 & 0.23 & 1.39 & 0.70 & 0.24 & 1.19 & 0.80 & 0.30 & 1.28 & 0.31 & 0.40 \\
 & MiMo-VL-7B~\cite{xiaomi2025mimovl} & 0.70 & 0.33 & 1.72 & 0.81 & 0.43 & 1.58 & 0.70 & 0.44 & 1.63 & 0.81 & 0.53 & 1.56 & 0.57 & 0.66 \\
\cmidrule(lr){2-16}
 & TimeLens-8B~\cite{zhang2025timelens} & 0.70 & \best{0.15} & 1.09 & 0.80 & 0.20 & 1.17 & 0.70 & 0.23 & 1.08 & 0.81 & 0.28 & 1.13 & 0.27 & 0.36 \\
 & TimeLens2-4B~\cite{zhu2026timelens2} & 0.70 & \best{0.15} & 1.02 & 0.80 & 0.19 & 1.05 & 0.70 & \best{0.22} & 1.02 & 0.80 & \best{0.26} & 1.04 & 0.25 & 0.35 \\
 & TimeLens2-8B~\cite{zhu2026timelens2} & 0.72 & \best{0.15} & 1.01 & 0.81 & \best{0.18} & 1.02 & 0.71 & \best{0.22} & 1.01 & 0.81 & \best{0.26} & 1.03 & \best{0.24} & \best{0.33} \\
\midrule
ActivityNet Captions & Qwen3-VL-4B~\cite{bai2025qwen3vl} & 0.70 & 0.38 & 1.37 & 0.80 & 0.50 & 1.35 & 0.70 & 0.47 & 1.20 & 0.80 & 0.58 & 1.15 & 0.66 & 0.71 \\
 & Qwen3-VL-8B~\cite{bai2025qwen3vl} & 0.70 & \best{0.35} & 1.30 & 0.80 & \best{0.44} & 1.43 & 0.70 & \best{0.45} & 1.21 & 0.80 & \best{0.54} & 1.22 & \best{0.62} & \best{0.70} \\
 & MiMo-VL-7B~\cite{xiaomi2025mimovl} & 0.70 & 0.49 & 1.57 & 0.80 & 0.59 & 1.27 & 0.70 & 0.59 & 1.30 & 0.80 & 0.67 & 1.25 & 0.71 & 0.78 \\
\cmidrule(lr){2-16}
 & TimeLens-8B~\cite{zhang2025timelens} & 0.70 & 0.38 & 1.31 & 0.80 & 0.50 & 1.29 & 0.70 & 0.49 & 1.17 & 0.80 & 0.60 & 1.13 & 0.66 & 0.74 \\
 & TimeLens2-4B~\cite{zhu2026timelens2} & 0.70 & 0.37 & 1.13 & 0.80 & 0.49 & 1.11 & 0.70 & 0.49 & 1.06 & 0.80 & 0.60 & 1.04 & 0.65 & 0.74 \\
 & TimeLens2-8B~\cite{zhu2026timelens2} & 0.70 & 0.36 & 1.13 & 0.80 & 0.46 & 1.14 & 0.70 & 0.49 & 1.07 & 0.80 & 0.59 & 1.05 & 0.63 & 0.74 \\
\bottomrule
\end{tabular}
\end{adjustbox}
\par\medskip\textbf{Comparison at tIoU 0.5 and 80\% requested success}\par\smallskip
\begin{adjustbox}{max width=\textwidth}
\begin{tabular}{@{}l ccc ccc ccc@{}}
\toprule
 & \multicolumn{3}{c}{Charades-STA} & \multicolumn{3}{c}{QVHighlights} & \multicolumn{3}{c}{ActivityNet Captions} \\
\cmidrule(lr){2-4}\cmidrule(lr){5-7}\cmidrule(lr){8-10}
\textbf{Model} & COVER & Greedy anchor & \texttt{CAPTURE} & COVER & Greedy anchor & \texttt{CAPTURE} & COVER & Greedy anchor & \texttt{CAPTURE} \\
\midrule
Qwen3-VL-4B & 0.59 & 0.38 & \best{0.34} & 0.71 & 0.37 & \best{0.23} & 0.70 & 0.58 & \best{0.50} \\
Qwen3-VL-8B & 0.54 & 0.34 & \best{0.33} & 0.71 & 0.37 & \best{0.23} & 0.70 & 0.57 & \best{0.44} \\
MiMo-VL-7B & 0.66 & 0.49 & \best{0.43} & 0.68 & \best{0.43} & \best{0.43} & 0.76 & 0.69 & \best{0.59} \\
TimeLens-8B & 0.57 & \best{0.35} & 0.39 & 0.64 & 0.25 & \best{0.20} & 0.70 & 0.58 & \best{0.50} \\
TimeLens2-4B & 0.59 & \best{0.37} & 0.38 & 0.58 & 0.22 & \best{0.19} & 0.69 & 0.59 & \best{0.49} \\
TimeLens2-8B & 0.58 & \best{0.37} & 0.38 & 0.55 & 0.22 & \best{0.18} & 0.69 & 0.58 & \best{0.46} \\
\bottomrule
\end{tabular}
\end{adjustbox}

\end{table*}

%% file: fig/model-watch.tex
\begin{figure}[t]
\centering
\includegraphics[width=\linewidth]{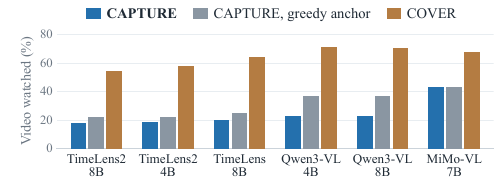}
\caption{\textbf{\texttt{CAPTURE} watches less across all six models.}
Mean QVHighlights watch time at 80\% requested success.
\texttt{CAPTURE} and its greedy-anchor variant guarantee a contained clip at IoU $\ge0.5$.
COVER~\cite{mohamed2026cover} guarantees ground-truth containment.}
\label{fig:model-watch}
\end{figure}

%% file: tables/set_methods.tex
\begin{table*}[t]
\caption{\textbf{\texttt{CAPTURE} watches least or ties in all six QVHighlights settings among methods with the same guarantee.}
Greedy has no guarantee. COVER guarantees containment averaged over query-level calibration sets.
The other rows are \texttt{CAPTURE} with a cheaper anchor or region in place of the exact anchor, with the same split-conformal calibration.
\Cref{tab:supp:watch,tab:supp:quality} compare Charades-STA and ActivityNet Captions in the supplement.}
\label{tab:methods}
\centering
\scriptsize
\setlength{\tabcolsep}{3.4pt}
\begin{adjustbox}{max width=\textwidth}
\begin{tabular}{@{}l cccc cccc cccc@{}}
\toprule
& \multicolumn{4}{c}{\textbf{Qwen3-VL-4B}} & \multicolumn{4}{c}{\textbf{TimeLens2-8B}} & \multicolumn{4}{c}{\textbf{Qwen3-VL-8B}} \\
\cmidrule(lr){2-5}\cmidrule(lr){6-9}\cmidrule(lr){10-13}
& \multicolumn{3}{c}{0.5, 80\%} & 0.7, 80\% & \multicolumn{3}{c}{0.5, 80\%} & 0.7, 80\% & \multicolumn{3}{c}{0.5, 80\%} & 0.7, 80\% \\
\cmidrule(lr){2-4}\cmidrule(lr){5-5}\cmidrule(lr){6-8}\cmidrule(lr){9-9}\cmidrule(lr){10-12}\cmidrule(lr){13-13}
\textbf{Method} & Hit & Watch$\downarrow$ & Seg. & Watch$\downarrow$ & Hit & Watch$\downarrow$ & Seg. & Watch$\downarrow$ & Hit & Watch$\downarrow$ & Seg. & Watch$\downarrow$ \\
\midrule
\gray{Greedy answer of the model} & \gray{0.75} & \gray{0.27} & \gray{1.00} & \gray{0.27} & \gray{0.88} & \gray{0.29} & \gray{1.00} & \gray{0.29} & \gray{0.74} & \gray{0.25} & \gray{1.00} & \gray{0.25} \\
\gray{COVER~\cite{mohamed2026cover}} & \gray{0.93} & \gray{0.71} & \gray{1.00} & \gray{0.71} & \gray{0.96} & \gray{0.55} & \gray{1.00} & \gray{0.55} & \gray{0.92} & \gray{0.71} & \gray{1.00} & \gray{0.71} \\
\midrule
\texttt{CAPTURE}, positional prior, no model & 0.83 & 0.77 & 2.78 & 0.87 & 0.83 & 0.77 & 2.78 & 0.87 & 0.83 & 0.77 & 2.78 & 0.87 \\
\texttt{CAPTURE}, greedy anchor & 0.81 & 0.37 & 1.00 & 0.46 & 0.83 & 0.22 & 1.00 & 0.29 & 0.81 & 0.37 & 1.00 & 0.46 \\
\texttt{CAPTURE}, MBR anchor at $\tau$ & 0.81 & 0.30 & 1.00 & 0.42 & 0.83 & 0.21 & 1.00 & 0.28 & 0.82 & \best{0.23} & 1.00 & 0.34 \\
\texttt{CAPTURE}, per-second threshold & 0.81 & 0.30 & 1.05 & 0.34 & 0.81 & 0.24 & 0.97 & 0.28 & 0.81 & 0.27 & 1.04 & \best{0.30} \\
\midrule
\rowcolor{oursblue!9} \textbf{\texttt{CAPTURE}}, exact anchor & 0.80 & \best{0.23} & 1.27 & \best{0.32} & 0.81 & \best{0.18} & 1.02 & \best{0.26} & 0.81 & \best{0.23} & 1.39 & \best{0.30} \\
\bottomrule
\end{tabular}
\end{adjustbox}

\end{table*}

%% file: fig/qualitative.tex
\begin{figure}[t]
\centering
\includegraphics[width=\linewidth]{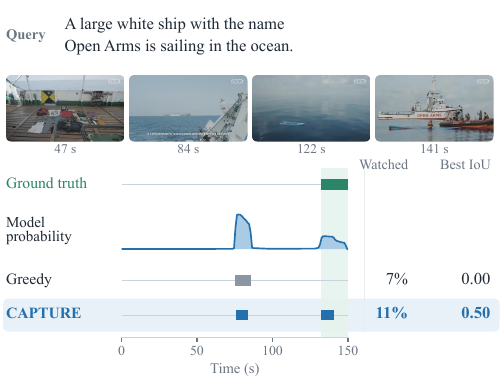}
\caption{\textbf{Two segments recover a moment missed by greedy.}
The blue curve gives TimeLens-8B's per-second moment probabilities.
Greedy selects the earlier peak. \texttt{CAPTURE} keeps both, leaving the gap unwatched.
Best IoU measures the best contained clip.
\texttt{CAPTURE} reaches 0.50 at 11\% watch time on QVHighlights with requested IoU 0.5 and 80\% success.}
\label{fig:qualitative}
\end{figure}

%% file: sec/6_conclusion.tex
\section{Limitations and conclusion}
\label{sec:conclusion}

\noindent\textbf{Limitations.}
New benchmarks or annotation protocols require fresh i.i.d. calibration videos.
Coverage is marginal, not guaranteed separately by query type, moment length or video duration.
Calibration keeps one query per video, discarding the other queries of each calibration video.
Viewers must still find the matching clip within selected footage.
Grounding-model errors limit how close a valid method can uniformly come to the oracle's watch duration, as the lower bound predicts.
For the 80\% requested success rate, the sufficient watch-duration condition requires overestimation plus underestimation below 0.12 (\cref{sec:supp:bound:practice}).
Mean overestimation alone reaches 0.26, so this condition does not certify short regions for today's grounders.
The condition is sufficient but conservative, so in practice we tune $q_0$ on fit videos, which keeps regions short while coverage still holds.

\noindent\textbf{Conclusion.}
For any fixed grounder, \texttt{CAPTURE} certifies that the selected footage contains a clip matching the moment at the requested tIoU and success rate.
The grounder identifies likely spans, and conformal calibration determines how much footage to return.
For tIoU of at least 0.5, its exact dynamic program guarantees minimum possible watch time at given model coverage.
Under model-error assumptions, our split-conformal analysis bounds watch duration against an oracle using the true conditional distribution.
A matching information-theoretic lower bound proves asymptotic minimax sharpness.
Faster algorithms preserve coverage on long videos.
Across three datasets and six models, mean watch duration is about $35\%$ below prior methods.
With \texttt{CAPTURE}, a person needs to watch much less of the video to find a matching clip with a prescribed probability.